\documentclass[11pt,a4paper]{article}

\usepackage[margin=2.5cm]{geometry}
\usepackage[super,comma,sort&compress]{natbib}
\usepackage{graphicx}
\usepackage{booktabs}
\usepackage{caption}
\usepackage{xcolor}
\usepackage{setspace}
\usepackage{lineno}
\usepackage{titlesec}
\usepackage[section]{placeins}
\usepackage[T1]{fontenc}
\usepackage{mathptmx}                 
\usepackage[scaled=0.92]{helvet}      
\usepackage{xurl}
\usepackage[hidelinks]{hyperref}

\definecolor{navy}{HTML}{245E84}
\definecolor{slate}{HTML}{4B5563}

\makeatletter
\renewcommand{\@biblabel}[1]{#1.}
\makeatother

\titleformat{\section}{\sffamily\bfseries\large}{}{0em}{}
\titleformat{\subsection}{\sffamily\bfseries\normalsize}{}{0em}{}
\titlespacing*{\section}{0pt}{18pt}{6pt}
\titlespacing*{\subsection}{0pt}{12pt}{4pt}

\DeclareCaptionLabelSeparator{bar}{ \textbar\ }
\usepackage{fancyhdr}
\fancypagestyle{titlefoot}{%
  \fancyhf{}%
  \fancyfoot[R]{\sffamily\footnotesize\itshape$^{*}$ Correspondence: xiaoyan.psych@gmail.com}%
}

\title{\vspace{-1.5cm}\sffamily\bfseries\LARGE Human-like moral judgments conceal divergent motive
attributions in large language models}
\author{%
  {\sffamily\bfseries Xiaoyan Wu$^{1*}$ and Jean-Claude Dreher$^{2}$}\\[8pt]
  {\sffamily\footnotesize$^{1}$Department of Adult Psychiatry and Psychotherapy, University of Zurich, Zurich, Switzerland}\\
  {\sffamily\footnotesize$^{2}$Institut des Sciences Cognitives Marc Jeannerod, CNRS UMR 5229, Lyon, France}}
\date{}

\begin{document}
\maketitle
\thispagestyle{titlefoot}

\section*{Abstract}
Large language models (LLMs) are used to simulate human participants in psychological research. We asked whether LLMs that reproduce human evaluations of a whistleblower’s moral character also reproduce the motive attributions that accompany them. Five LLMs and two human samples ($N = 125$ and $N = 742$) evaluated a physician
who either remained silent about fraudulent billing or reported it to a hospital, regulator, or
newspaper. Models reproduced the human ranking of the physician's moral character but portrayed
whistleblowers as more helpful, less self-interested, and less hostile. In four of five models,
competitive motives were less strongly associated with moral-character judgments. Model ratings
changed little when prompts reproduced the narratives and demographic profiles of both human
samples, although this comparison cannot isolate a perspective effect. Thus, agreement in average
ratings can conceal differences in attributed motives, relationships among judgments, and
sensitivity to context. Validating LLMs as simulated participants therefore requires testing
psychologically informative response patterns, not average agreement alone.

\vspace{0.3cm}
\noindent\textbf{Keywords:} large language models; simulated participants; moral judgment; motive
attribution; whistleblowing

\section*{Introduction}

Large language models (LLMs) are increasingly being used to approximate human responses in
psychological and social-science research. Prompted models can forecast aggregate treatment effects
across experiments \citep{Ashokkumar2026}, interview-grounded agents can approximate responses from
particular individuals \citep{Park2024agents}, and models trained on trial-level behavioural data
can predict choices across diverse cognitive tasks \citep{BinzCentaur2025}. These results establish
predictive value and have encouraged discussion of LLMs as scalable substitutes for human samples
\citep{Dillion2023}. However, a model can reproduce a human answer without reproducing the
psychological structure that makes the answer scientifically informative. Forecasting an average
experimental effect, reproducing an individual trajectory and representing the relationships among
psychological constructs are different validation problems.

This distinction is especially important because recent work shows that human-like performance can
coexist with substantial departures from human response patterns. Across 156 experiments in
psychology and management, LLMs reproduced main effects more reliably than interactions and often
magnified effect sizes \citep{Cui2025}. In a preregistered moral-judgment study, model and human
averages were almost perfectly correlated even though model ratings differed from human means on
most scenarios and occupied a much narrower response range \citep{Grizzard2025}. A population-level
benchmark spanning 15 LLMs and seven major social surveys similarly found compressed distributions
and poor reproduction of longitudinal patterns \citep{Xie2026}. Model responses can also vary with
seemingly minor prompt changes and model updates \citep{bisbee2024synthetic}, while demographic
prompting may flatten or misrepresent the groups it is intended to simulate \citep{WangIdentity2025}.
Current LLMs therefore appear better at recovering some average regularities than at reproducing
human variation and conditional relationships.

Most human-LLM comparisons have focused on final choices, mean ratings, or treatment effects. Yet in
many psychological experiments, the interpretation of those outcomes depends on how different
measures relate to one another. Groups can reach different decisions despite similar beliefs
because they value reciprocity differently \citep{WuAdolescents2026}, and, conversely, two
respondent groups may reach similar judgments while differing in the motives, beliefs, or values
associated with those judgments. Nor can this problem be resolved simply by asking a model to
explain its answer, because generated rationales need not faithfully reflect the information that
shaped the output \citep{Turpin2023,ChenCoT2025}. A stronger comparison therefore requires
examining not only whether humans and models give similar answers, but also whether theoretically
relevant measures show similar profiles and relationships, and whether they respond similarly to
changes in context.

Moral judgments of whistleblowers provide a useful test of this question. Whistleblowing is morally
ambiguous: an observer may regard disclosure as dutiful and socially beneficial while also
attributing disloyalty, self-interest, or hostility \citep{NearMiceli1985,MesmerMagnus2005}.
Judgments of the same act can depend on the motives attributed to the actor, consistent with
attributional accounts of intention, blame, and moral evaluation
\citep{Weiner1985,MalleKnobe1997,GrayYoungWaytz2012}. In the human benchmark used here, respondents
evaluated a physician who either remained silent about fraudulent billing or disclosed it
internally, to an external authority, or to the press \citep{Brotzeller2025}. These disclosure
channels altered duty-based, prosocial, individualistic, and competitive motive ratings, and those
ratings were associated with judgments of moral character. Because several motives can contribute
to moral evaluation at the same time \citep{wu2024motive,WuChildren2026}, a model could reproduce
the ordering of character judgments while construing the actor's motives differently from human
observers.

The published dataset also contains two independently recruited human samples that received
closely related versions of the scenario. In the first-hand sample, the events were described as a
recent occurrence within the respondent's own team; in the second-hand sample, respondents were
told that they had heard about the events after they occurred. Psychological distance and observer
perspective can influence social attribution \citep{TropeLiberman2010,Eyal2008,Malle2006}.
However, framing, recruitment, and sample composition differed simultaneously, so the contrast
between these samples cannot identify a causal effect of perspective. We therefore use it only as
an exploratory test of whether LLM responses distinguish between the two complete sample
specifications. Matching, adjustment, and weighting can address differences between simulated
persona rosters, but they cannot remove the confounding in the original human comparison.

\begin{figure}[!htb]
\centering
\includegraphics[width=1\linewidth]{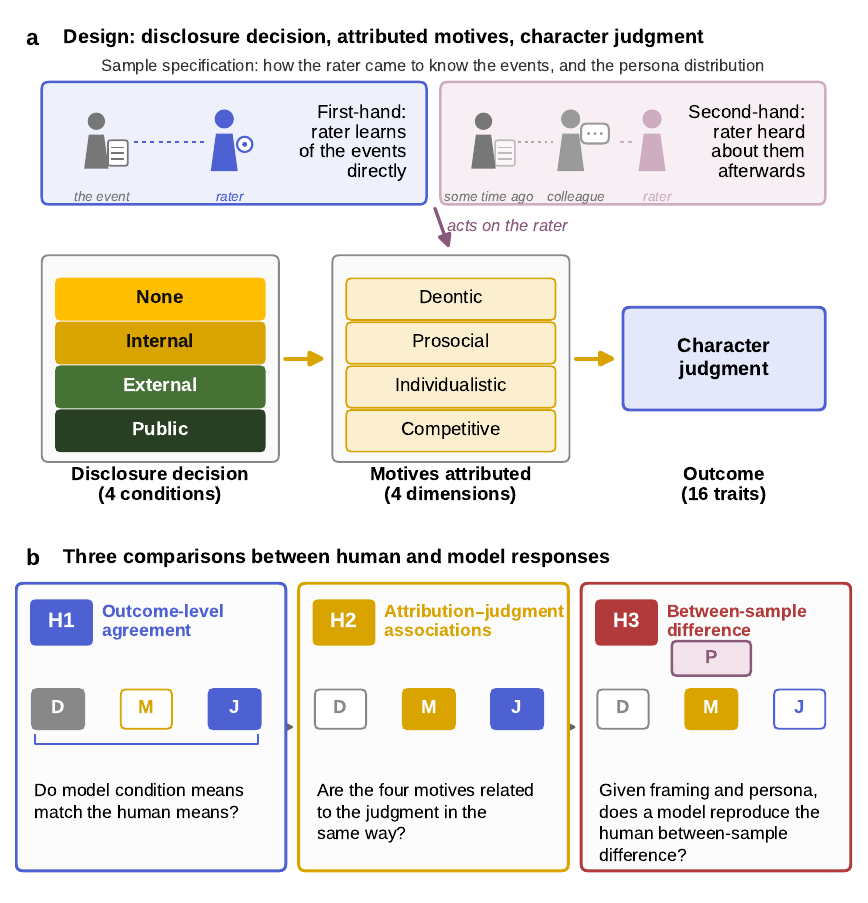}
\caption{\textbf{Analytical framework for evaluating human-LLM correspondence.} \textbf{a}, The
benchmark design: the physician's disclosure decision (four between-subjects conditions) shapes
the motives a rater attributes to the physician (four dimensions), which are related to the rater's
judgment of the physician's moral character (16 traits). The sample specification, narrative
framing together with the persona distribution, distinguishes the two human samples ($N = 125$
first-hand, $N = 742$ second-hand; Brotzeller et al.) and the two model runs (five models, 200
demographically matched personas per condition and specification). In the first-hand framing the
events are narrated to the rater directly, as a recent occurrence in their own team (grey figure
with the billing form); in the second-hand framing the rater has heard about the same events some
time after they occurred. \textbf{b}, The three
comparisons, each with a schematic of the chain (D, disclosure decision; M, motive attributions;
J, character judgment; P, sample specification) in which filled elements mark what is compared
between humans and models. H1 compares the endpoints only. H2 compares associations among ratings
produced together; it does not identify how either humans or models reached a judgment. H3 asks
whether a model given both the framing and the matched persona distribution reproduces the
between-sample difference; it does not require that difference to be causally identified. Arrows
indicate the order of measurement and comparison and do not imply mediation or internal causal
pathways. The human samples were not jointly randomized to narrative framing.}
\label{fig:framework}
\end{figure}

We compared humans and five closed- and open-weight LLMs at three levels
(Fig.~\ref{fig:framework}). First, we asked whether model condition means reproduced the human
pattern across disclosure conditions (H1). Second, we tested whether the four motive attributions
were related to moral-character judgments in the same way in humans and models (H2). These
associations describe the organization of co-produced ratings and are not evidence of mediation or
internal computation. Third, we asked whether models given the two published sample specifications
reproduced the observed between-sample difference (H3). The models broadly recovered selected
outcome-level patterns, but they portrayed disclosure as more prosocial and less hostile than
humans did, and competitive motive ratings were less strongly related to moral-character judgments
in four of five models. Thus, similarity in moral judgments did not imply similarity in the motive
attributions associated with those judgments.

\section*{Results}

\subsection*{Models reproduce selected outcome-level patterns}

Across the four disclosure conditions, model and human condition means (Supplementary
Tables 2--4) were strongly correlated for
moral-character judgments in the first-hand comparison ($r = 0.90$ to $0.95$) and
somewhat more variably in the second-hand comparison ($0.75$ to $0.94$). Duty-based
(deontic) attributions showed similarly high correlations ($r = 0.84$ to $0.97$). These coefficients
summarize only four condition means and should not be treated as sufficient evidence of
correspondence. Bootstrap intervals were comparatively narrow for character and deontic ratings but
wide for the other motive dimensions (Supplementary Table~5). We emphasize contrasts and
absolute levels rather than correlations alone.

Most models reproduced the broad human pattern that disclosure was judged more favourably than
silence, with internal and external reporting generally receiving the highest character ratings. The
match was not exact. For example, Llama-3.1-8B-Instruct rated silence and public disclosure almost identically in
the first-hand comparison ($4.09$ versus $4.08$; contrast $-0.01$, 95\% CI $[-0.18,
0.15]$) despite a profile correlation of $r = 0.92$. High correlation can thus coexist with loss of
a contrast central to the scenario.

\begin{figure}[!htb]
\centering
\includegraphics[width=\linewidth]{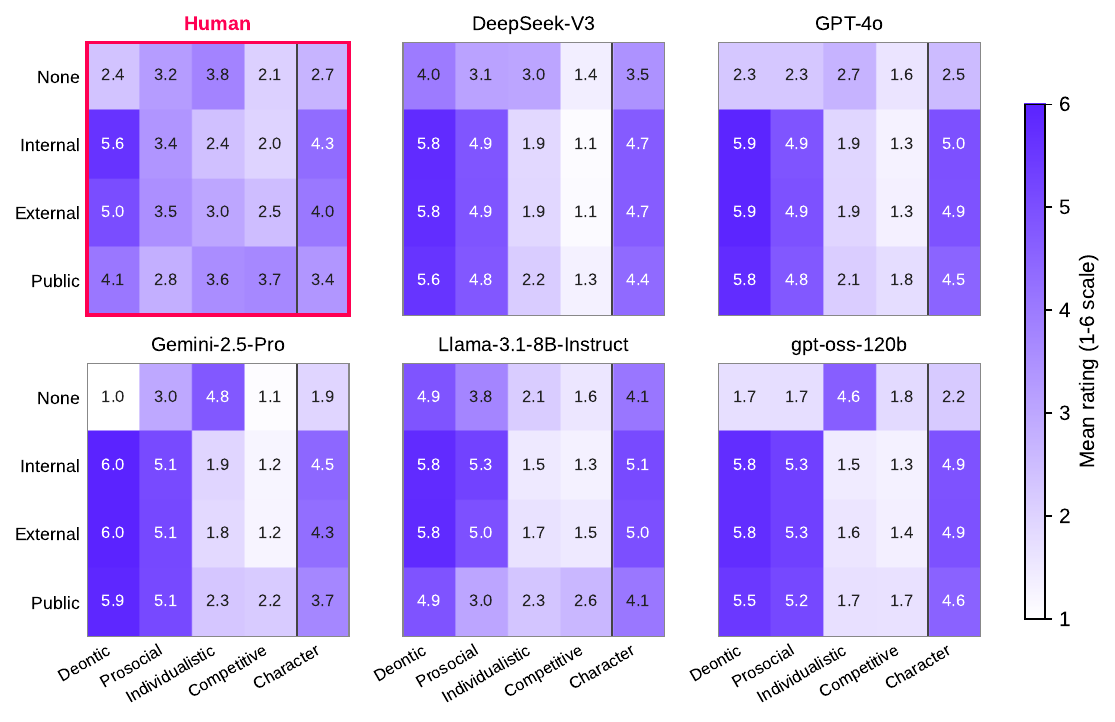}
\caption{\textbf{Condition-by-outcome profiles.} Mean rating (1--6 scale) for the human
first-hand sample (top left, outlined in red) and for each of the five models, on a shared
colour scale. Rows are the four disclosure conditions (None, no whistleblowing); the first four columns are the motive
dimensions and the fifth is the character judgment. Models reproduce the human rise in deontic
attributions and in the character judgment once a disclosure occurs, but every model's prosocial
column also rises sharply where the human column stays nearly flat.}
\label{fig:profiles}
\end{figure}

Two comparisons separated model and human ratings (Table~\ref{tab:contrasts}; Fig.~\ref{fig:profiles}).
Moving from silence to any disclosure increased human deontic attributions by $2.55$ scale points
(95\% CI $[2.03, 3.07]$) and character judgments by $1.23$ points $[0.86, 1.60]$; every model
reproduced both shifts in direction, although not always in magnitude. Human prosocial attribution
did not change ($-0.01$, $[-0.50, 0.48]$), whereas every model interpreted disclosure as more
prosocial, with increases of $0.65$ to $3.52$ points. Competitive attribution increased in humans
($+0.66$, $[0.28, 1.05]$) but showed small or inconsistent changes across models ($-0.38$ to
$+0.49$).

\begin{table}[!htb]
\centering
\caption{Change in mean rating from no whistleblowing to any disclosure (first-hand
materials, scale points). Positive values indicate higher ratings when a disclosure occurred. Human
estimates are shown with 95\% confidence intervals.}
\label{tab:contrasts}
\small
\setlength{\tabcolsep}{4pt}
\begin{tabular}{@{}lccccc@{}}
\toprule
 & \shortstack{Character\\judgment} & Deontic & Prosocial & Individualistic & Competitive \\
\midrule
Human            & $+1.23$ & $+2.55$ & $-0.01$ & $-0.80$ & $+0.66$ \\
\quad 95\% CI    & {\scriptsize$[0.86, 1.60]$} & {\scriptsize$[2.03, 3.07]$}
                 & {\scriptsize$[-0.50, 0.48]$} & {\scriptsize$[-1.22, -0.38]$}
                 & {\scriptsize$[0.28, 1.05]$} \\
\addlinespace
DeepSeek-V3      & $+1.12$ & $+1.74$ & $+1.74$ & $-1.06$ & $-0.20$ \\
GPT-4o           & $+2.27$ & $+3.62$ & $+2.60$ & $-0.73$ & $-0.15$ \\
Gemini-2.5-Pro   & $+2.21$ & $+4.95$ & $+2.15$ & $-2.76$ & $+0.49$ \\
Llama-3.1-8B & $+0.62$ & $+0.62$ & $+0.65$ & $-0.31$ & $+0.26$ \\
gpt-oss-120b     & $+2.61$ & $+3.93$ & $+3.52$ & $-3.07$ & $-0.38$ \\
\bottomrule
\end{tabular}
\end{table}

Averaged across the design, models assigned prosocial motives $1.13$ points more strongly than
humans and competitive motives $1.07$ points less strongly. They also assigned self-interested
motives $0.91$ points less strongly and deontic motives $0.70$ points more strongly.
Root-mean-square deviations from human cell means were $1.1$ to $1.4$ points on the six-point scale
for every model in both comparisons, and the pattern of cell-level deviations was closely similar
under the two sample specifications (Supplementary Figs.~1 and~2). Models broadly agreed with humans about which disclosure
choices merited more favourable character judgments while construing the discloser as more
altruistic, less self-interested, and less hostile.

\FloatBarrier
\subsection*{Attribution-judgment associations differ most for competitive motives}

\begin{figure}[!htb]
\centering
\includegraphics[width=\linewidth]{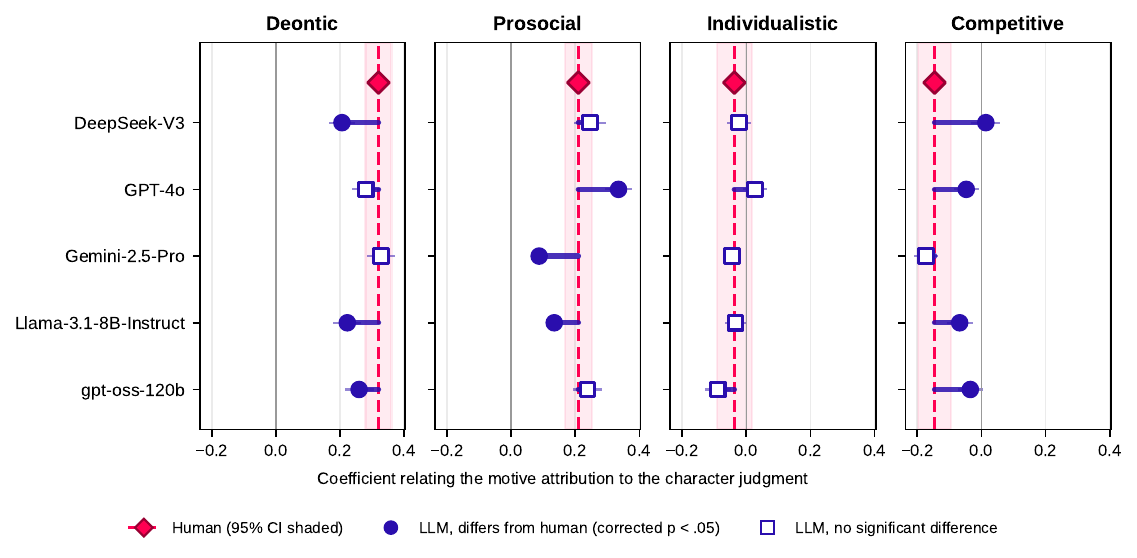}
\caption{\textbf{Attribution-judgment associations.} The human coefficient relating each motive
attribution to the character judgment (red diamond and dashed line, 95\% CI shaded) versus each
model's coefficient (marker, with 95\% CI), controlling for disclosure condition and sample
specification. Filled circles mark models whose coefficient differs from the human value at
corrected $P < 0.05$; open squares mark no significant difference. Coefficients describe associations
among co-produced ratings and are not mediation paths.}
\label{fig:bpath}
\end{figure}

We next compared the association of the four motive ratings with the character judgment in humans
and models, controlling for disclosure condition and sample specification (Fig.~\ref{fig:bpath}).
Because motives and character were rated in the same response, these
coefficients describe conditional associations among bounded measures. They do not identify a causal
pathway or the computations that generated either response.

The motive-by-respondent-type interaction was significant for competitive attribution in four of
five models; Gemini-2.5-Pro was the exception ($P = 0.051$). Human respondents judged an actor they
saw as competitively motivated, seeking to harm a colleague, less favourably (coefficients from
$-0.13$ to $-0.15$), whereas competitive ratings had weaker associations with character in most
models (coefficients from $-0.07$ to $+0.01$). Deontic attribution differed between humans and three
models, in each case because the model coefficient was smaller. Prosocial attribution showed a mixed
pattern, whereas individualistic attribution was comparably associated with character across
respondent types. The pattern survived within-model false-discovery-rate correction. In
sample-size-matched subsampling, the competitive interaction remained significant in more than 99\%
of draws for DeepSeek-V3 and gpt-oss-120b, 75\% for Llama-3.1-8B-Instruct, and 59\% for GPT-4o
(Supplementary Table~9).

These results indicate that outcome similarity did not extend uniformly to how the ratings related
to one another. They should still be interpreted with the scale properties in mind: model
ratings were more concentrated near the scale ceiling, and range restriction can affect regression
coefficients. Gemini-2.5-Pro is especially uncertain because its deontic composite had zero variance
in two disclosure conditions.

\FloatBarrier
\subsection*{Exploratory comparison of the two human-sample specifications}

The two published human samples differ in narrative framing, recruitment, age, and student
composition. The contrast is not a causal estimate of perspective. We ask a narrower,
exploratory question: when each model is given the framing and matched persona distribution
corresponding to each human sample, does it reproduce a difference of the observed magnitude?

Controlling for disclosure condition, the second-hand human sample attributed $0.34$
fewer scale points of deontic motive than the first-hand sample (95\% CI $[-0.55,
-0.13]$, $P = 0.002$). The largest differences occurred for internal disclosure ($d = -0.65$) and
public disclosure ($d = -0.41$); the difference was smaller for external disclosure ($d = -0.16$)
and absent when the physician remained silent ($d = -0.03$), where deontic ratings were near the
floor in both samples. Individualistic attribution differed only in the no-disclosure condition
($-0.52$ points, $d = -0.48$); prosocial and competitive attributions did not differ reliably
between samples (Supplementary Table~8).

\begin{figure}[!htb]
\centering
\includegraphics[width=\linewidth]{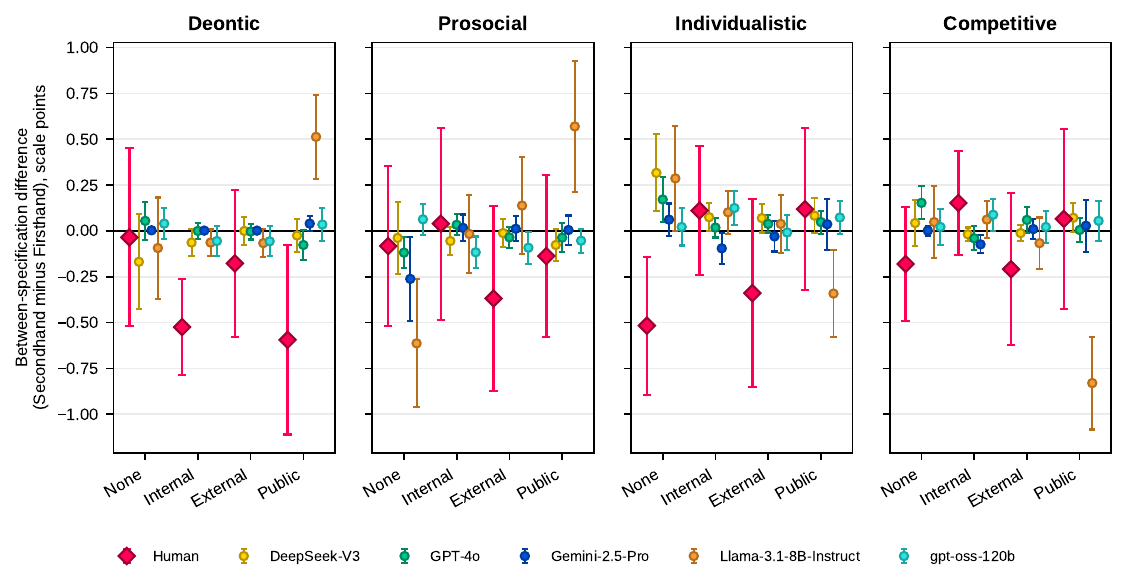}
\caption{\textbf{Exploratory reproduction of the between-sample difference.} Difference in mean
motive rating between the second-hand and first-hand specifications (scale points), by
disclosure condition (None, no whistleblowing), for the human samples (red diamonds) and each of the five models (coloured
circles), with 95\% confidence intervals. The human contrast is cross-study and is not a causal
estimate of narrative perspective.}
\label{fig:shift}
\end{figure}

Every model was run under both specifications. Model estimates of the deontic difference ranged from
$-0.06$ to $+0.07$ scale points, compared with $-0.34$ in the human samples (Fig.~\ref{fig:shift}).
Using an a priori equivalence bound of $0.20$ scale points, deontic differences were statistically
equivalent to zero in all five models (all $P_\mathrm{TOST} \leq 0.006$). These tests establish small
differences between the two model specifications; they do not establish that the human difference
was caused by perspective.

The simulated persona rosters differed between model runs, so we conducted three model-side
sensitivity analyses. Adjustment for persona age, gender, and student status yielded estimates from
$-0.17$ to $+0.04$. Restricting the analysis to age-by-gender-by-student-by-condition cells
represented in both framing conditions (retaining 85\% of responses) yielded estimates from $-0.09$
to $+0.06$. Stabilized inverse-probability weighting yielded estimates from $-0.15$ to $+0.04$,
although residual age imbalance remained (Supplementary Tables~11 and~12). Across approaches, model
differences remained substantially closer to zero than the observed human difference. Because the
human contrast remains confounded, this result is evidence of weak sensitivity to the combined
specifications tested here, not evidence that LLMs lack a human perspective effect.

\FloatBarrier
\subsection*{Model ratings show limited sensitivity to simulated respondent characteristics}

Persona attributes were randomly sampled within each model's generation pipeline, allowing their
associations with model ratings to be estimated without the between-study confounding present in
the human comparison. Across the 15.7-year difference in mean age between human samples, the
implied change in model ratings was at most $0.22$ scale points and typically below $0.10$;
student-status associations were similarly small (Supplementary Table~10). Participant-level
demographics were unavailable in the published human data, so these estimates cannot be compared
directly with corresponding human associations. They describe the tested models and prompts:
specifying limited demographic characteristics produced little differentiation in their ratings.

\subsection*{Measurement properties}

Internal consistency was comparable across respondent types. Cronbach's $\alpha$ ranged from
$0.82$ to $0.97$ in the human samples and from $0.82$ to $0.99$ across models and scales. Even so, model responses were more concentrated at the maximum. The proportion of responses assigning
6 to all three deontic items ranged from 45\% to 71\% across models; for Gemini-2.5-Pro, the
deontic composite had zero variance in the internal and external conditions. Excluding those
conditions changed some of that model's attribution-judgment comparisons (Supplementary
Table~9). We therefore interpret its coefficient differences cautiously and treat high internal
consistency as insufficient evidence of human-like measurement behaviour.

\section*{Discussion}

Five LLMs evaluated the same whistleblowing scenarios as two published human samples. Most models
recovered the broad ordering of moral-character judgments and duty-based motive ratings across
disclosure channels, which is the level at which human-LLM agreement is often reported. The
agreement did not extend to all aspects of the response. Every model treated disclosure as evidence
of a stronger desire to help others even though human prosocial ratings were essentially unchanged,
and models generally attributed less self-interest and hostility. Competitive (hostile) motive
ratings were also less strongly associated with character judgments in four of five models. Similar
moral judgments therefore coexisted with different motive profiles and different relationships
among the ratings.

This difference matters because whistleblowing is morally ambiguous. Human observers may see the
same disclosure as dutiful or socially beneficial while also suspecting hostility, betrayal, or
personal advantage \citep{NearMiceli1985,MesmerMagnus2005,Brotzeller2025}. The models preserved the
positive relation between disclosure and moral character but expressed less of this ambivalence.
This is consistent with reports that LLMs can amplify particular moral tendencies \citep{Cheung2025}
and can generate advice perceived as highly virtuous \citep{Dillion2025}. Why this occurs is not
established by the present data. Training distributions, post-training procedures, prompt
interpretation, and differences in scale use are all possible contributors. Our results therefore
concern the pattern of model judgments rather than the mechanisms that produced them.

The findings extend earlier comparisons of human and model moral judgments. Grizzard et al.\ showed
that very high correlations between human and model ratings can coexist with substantial
differences in absolute ratings and response ranges \citep{Grizzard2025}. Here, the discrepancy
goes beyond calibration. Disclosure produced almost no change in prosocial attribution in humans but
increased it in every LLM, whereas competitive attribution increased in humans but changed little
or inconsistently across models. In addition, competitive attribution was less strongly associated
with moral-character judgments in four of the five models. These differences would largely remain
even if model ratings were rescaled to better match human means, because the disagreement concerns
which motives are associated with the act and how those motives relate to moral evaluation.

The attribution-judgment analyses reinforce this point, while remaining correlational. Because
motive and character ratings were generated within the same response, these coefficients do not
establish that humans or models first inferred a motive and then used it to form a moral judgment.
They show instead that the observed ratings were related differently across respondent types. This
complements large-scale evidence that LLMs reproduce main effects more reliably than interactions
and may exaggerate effect sizes \citep{Ashokkumar2026,Cui2025}. Reproducing an average effect is
therefore not sufficient when the psychological question concerns how different components of
judgment are related.

The comparison between the two human samples provides a more tentative test of contextual
sensitivity. Deontic attribution differed by about one-third of a scale point between the human
samples, whereas the corresponding model specifications differed by less than one-tenth of a point
in the unadjusted analysis. Because the human samples also differed in recruitment, demographics,
and other unmeasured study features, this contrast cannot show that humans, but not models, respond
to first-hand versus second-hand perspective. It shows only that the tested models changed little
between two complete specifications associated with distinguishable human samples. A causal
human-LLM comparison would require a new experiment in which narrative perspective is randomized
within the same recruitment frame and crossed with disclosure condition.

Related work on theory-of-mind benchmarks shows why such tests are useful: model performance on
familiar tasks can deteriorate after small changes that preserve the underlying problem, and
estimates of competence can depend strongly on task construction
\citep{Ullman2023,Strachan2024,Kosinski2024}. These studies do not explain the present findings,
but they illustrate the risk of treating success on a recognizable vignette as evidence of broader
contextual generalization.

Our results suggest several practical checks before LLM responses are treated as substitutes for
human data. Matching the direction or ordering of an experimental effect is not enough. Researchers
should also compare absolute response levels and distributions, test whether theoretically
important variables are related in similar ways, and examine responses to manipulations known to
affect human judgment. Tests using paraphrased or novel scenarios, other languages, preregistered
prompts, and out-of-sample prediction would help distinguish robust correspondence from familiarity
with a particular vignette. These checks are compatible with proposals to evaluate AI systems using
task-specific profiles rather than a single measure of human likeness \citep{HernandezOrallo2026}
and with mixed designs in which model outputs complement rather than replace human observations
\citep{Abdurahman2024,Broska2025}.

The weak differentiation produced by demographic persona cues also echoes concerns about using
LLMs to represent population heterogeneity. Previous work has shown that synthetic samples can
compress human variation \citep{Xie2026}, flatten or misrepresent demographic groups
\citep{WangIdentity2025}, and vary with analytic choices \citep{Cummins2026,Lin2025fallacies}.
Reviews of synthetic social agents have identified related problems, including cultural and
linguistic bias, atemporality, and disembodiment \citep{Kozlowski2025}. In the present data, age,
gender, and student-status prompts produced little differentiation, while several model response
distributions were concentrated near the top of the scale. Drawing more responses from the same
model therefore provides repeated samples from one system rather than additional independent
respondents.

Several limitations qualify these conclusions. The comparison between the two human samples is
observational and cross-study, the analyses were not preregistered, and the joint human-sample
contrast was selected after inspection of the benchmark data. It cannot support a causal claim about
narrative perspective. The study also relies on a single English-language whistleblowing paradigm and five selected
LLMs, so the findings apply to the model versions and prompts tested here. Replication
across other scenarios, languages, cultures, and morally ambiguous actions will be needed to
determine how general the observed dissociation is.

The attribution-judgment analyses also concern associations between bounded ratings produced within
the same response rather than a causal sequence from motive inference to moral evaluation. Ceiling
effects, restricted response ranges, and multicollinearity may affect the estimated coefficients, and
alternative ordinal or predictive analyses would provide useful robustness checks. Finally,
simulated personas varied only in age, gender, and student status, while participant-level
demographic data were unavailable for estimating corresponding moderation in the human samples.

The present findings show that agreement in moral judgments does not necessarily extend to the
motives associated with those judgments. LLMs may therefore be useful for predicting some aggregate
responses without providing an equivalent model of human psychological judgment. Establishing when
simulated participants are informative will require testing not only whether they reproduce what
people judge, but also whether they preserve the psychological relationships that give those
judgments their scientific meaning.

\section*{Methods}

\subsection*{Human benchmark samples}

Two human samples were drawn from the published dataset of Brotzeller et al.\ \citep{Brotzeller2025},
corresponding to that paper's Studies 2 and 3. We refer to the studies by their narrative framing.
The first-hand sample ($N = 125$) was recruited through a university mailing list and a
popular-psychology magazine website; the scenario narrated the events to
respondents directly, as a recent occurrence in their own team. The second-hand sample ($N = 742$) was recruited through an online
panel; the scenario told respondents that they had heard about the same
events some time after they occurred.
Demographic composition was: first-hand, age $M = 33.69$, $SD = 11.76$, 77.6\% female, 30.4\%
students; second-hand, age $M = 49.43$, $SD = 15.06$, 64.3\% female, 5.7\% students. Applying the
original screening criteria, one use-me item, a full attention-check score, and two correct
comprehension checks, reproduced the published sample sizes.

The original investigators described the wording changes between Studies 2 and 3 as minor and did
not design study membership as a randomized perspective manipulation \citep{Brotzeller2025}. Because
the samples were separately recruited and differ demographically, their contrast reflects an unknown
combination of framing, composition, recruitment, and other study-level differences. We therefore
call it a between-sample difference. The present study reanalysed de-identified data available from
the authors' public repository and collected no new human data. Ethical procedures for the original
data collection are reported in the source publication \citep{Brotzeller2025}.

\subsection*{Materials and measures}

Scenario materials were taken from Brotzeller et al.\ \citep{Brotzeller2025}. A physician, Dr
Schmitt, is described as fraudulently billing a health insurer for services not rendered. A
colleague, Dr Bauer, learns of the misconduct and either does not report it, reports it internally
to the hospital compliance department, reports it to an external regulatory authority, or discloses
it to a national newspaper. The four disclosure conditions were presented between participants. The
perspective versions differ in the framing sentence and the tense and viewpoint markers it entails.

Respondents rated 12 motive-attribution items, three each for the deontic, prosocial,
individualistic, and competitive dimensions, and 16 moral-character traits on the original six-point
scale. Item order was randomized independently for each respondent. Complete wording is provided in
Supplementary Table~1.

\subsection*{Simulated respondents}

Five LLMs served as simulated respondent pools: DeepSeek-V3, GPT-4o (\texttt{gpt-4o-2024-08-06}),
Gemini-2.5-Pro, Llama-3.1-8B-Instruct, and gpt-oss-120b. Access routes, prompts, and output schemas
are reported in Supplementary Methods~1.4. For each of the four disclosure conditions, each model
was queried independently until 200 valid simulated responses were obtained for each sample
specification (800 responses per model per specification). For each query, a persona comprising age, gender, and student status was sampled to
approximate the demographic composition of the corresponding human sample; the genders of the two
scenario characters were independently randomized. Sampling temperature was 0.8. Responses were
constrained to structured JSON containing numeric ratings and no free-text justification. Between
151 and 200 of the 200 responses in each cell were distinct 28-item vectors (Supplementary
Table~7).

Because personas were sampled to correspond to each human sample, the two model runs differed in
both framing and persona distribution (mean age 34.7 versus 50.4 years; 29.5\% versus 4.5\%
students). Their unadjusted contrast compares the specifications as a whole.
Persona-adjusted, matched, and weighted estimates are reported as sensitivity analyses; none can
remove confounding from the human between-study contrast.

\subsection*{Statistical analysis}

All regression-based tests used heteroskedasticity-consistent HC3 standard errors
\citep{mackinnon1985some}. Analyses were not preregistered.

\textbf{Outcome-level agreement.} For each sample specification, we calculated Pearson correlations
between human and model condition means separately for each outcome, together with root-mean-square
deviation from human cell means. Because each correlation contains only four condition means, we
obtained percentile intervals by bootstrapping respondents within condition 2,000 times. Substantive
interpretation emphasizes the disclosure contrasts in Table~\ref{tab:contrasts} and absolute levels.
Tucker's congruence coefficients from an earlier analysis are retained only in Supplementary
Table~6 because bounded positive ratings made them uniformly high even for substantively mismatched
profiles.

\textbf{Attribution-judgment associations.} We regressed moral-character judgment on the four
motive attributions, respondent type, and their interactions, fitting each model against the pooled
human data and controlling for disclosure condition and sample specification. The focal terms were
motive-by-respondent-type interactions. Because motives and character were rated within the same
response, these are conditional associations among co-produced measures; they are not mediation
estimates or evidence about internal processing. Sensitivity analyses for unequal sample size and
zero-variance cells are reported in Supplementary Table~9.

\textbf{Exploratory reproduction of the between-sample difference.} We estimated the second-hand minus first-hand difference,
controlling for disclosure condition, separately in the human data and within each model. Human
estimates are between-study associations, not causal perspective effects. Equivalence was assessed
using two one-sided tests with an a priori bound of $\pm 0.20$ scale points, chosen as the smallest
difference of interest on a six-point scale. Tests using the observed human difference as a
descriptive, data-derived bound are reported in the Supplementary Information.

Three analyses assessed sensitivity to the different simulated persona rosters. First, estimates
were adjusted for persona age, gender, and student status. Second, analyses were restricted to a
matched set containing age-band (seven bands) by gender by student-status by disclosure-condition by
model cells represented under both framings, retaining 6,777 of 8,000 responses per model set.
Third, stabilized inverse-probability weights were estimated from persona age, age squared, gender,
and student status and trimmed at the 1st and 99th percentiles. Weighting reduced the standardized
age difference from approximately 1.2 to 0.3 but did not eliminate it; weighted results are
therefore presented as sensitivity estimates rather than identified effects.

\textbf{Persona sensitivity and multiple comparisons.} Within each model pipeline, each outcome was
regressed on persona age, gender, and student status while controlling for disclosure condition and
framing. The age coefficient was expressed as the change implied by the 15.7-year difference between
human-sample means. For attribution-judgment interactions and between-specification tests,
Benjamini-Hochberg correction was applied within each model's four motive-dimension tests, with
adjusted and unadjusted values reported.

\section*{Data availability}
Human benchmark data are available from the original authors' OSF repository:
\url{https://osf.io/q4rsn}. Model-generated data are available at
\url{https://github.com/xiaoyanwu2024/AI_MoralReasoning}.

\section*{Code availability}
Data-collection scripts, analysis code, and prompt templates are available at
\url{https://github.com/xiaoyanwu2024/AI_MoralReasoning}.

\section*{Acknowledgements}
X.W. acknowledges support from the University of Zurich Postdoc Grant 2026. The authors thank Fan
Shi for technical assistance.

\section*{Author contributions}
X.W.: conceptualization, methodology, software, formal analysis, data curation, visualization,
writing (original draft), and writing (review and editing). J.-C.D.: conceptualization,
supervision, and writing (review and editing).

\section*{Competing interests}
The authors declare no competing interests.

\section*{Declaration of generative AI use}
The authors used Claude (Anthropic) to assist with language editing, drafting, and statistical
coding. The authors reviewed and edited all content and take full responsibility for the
publication.

\bibliographystyle{unsrtnat}
\bibliography{refs}

\end{document}


\begin{center}
{\sffamily\large\bfseries Supplementary Information}\\[6pt]
{\normalsize for: Human-like moral judgments conceal divergent motive attributions in large
language models}\\[6pt]
{\normalsize Xiaoyan Wu$^{1}$ and Jean-Claude Dreher$^{2}$}\\[3pt]
{\small $^{1}$Department of Adult Psychiatry and Psychotherapy, University of Zurich, Zurich,
Switzerland\\ $^{2}$Institut des Sciences Cognitives Marc Jeannerod, CNRS UMR 5229, Lyon, France}
\end{center}

\vspace{0.6cm}

\section{Supplementary Methods}\label{sec:methods}

\subsection{Prompt templates and model access}\label{sec:S1}

\subsubsection{System prompt (persona instruction)}
Bracketed fields were filled per query from the persona sampled for that respondent.

\begin{quote}
\ttfamily\small
You are participating in a psychology survey about how people interpret others' behavior in
workplace scenarios.\\[4pt]
You are answering as a \{age\}-year-old \{gender\} from Germany. \{student\_clause\}You are a
general member of the public taking part in an online survey, not an expert in ethics, law, or
psychology.\\[4pt]
Read the scenario carefully, then answer every statement below as this person would, based only on
the information given. There are no right or wrong answers; give your genuine personal impression.
Respond ONLY in the JSON format specified at the end. Do not add any explanation outside the JSON.
\end{quote}

\subsubsection{Scenario materials}
The scenario text combined a fixed introduction, which differed between the two perspective
conditions, with one of four condition-specific endings.

\paragraph{First-hand introduction.}
\begin{quote}
\ttfamily\small
You are a physician at the hospital "Klinik am See". You have already been working with your team
in this hospital for several years. Together with your team, you advise people who are ill on a
daily basis, make diagnoses and carry out treatments.\\[4pt]
The other day \{perpetrator\_ref\} from your team, Dr Schmitt, billed the health insurance company
for more services than were actually provided for some patients. For example, services were billed
by Dr Schmitt without \{perpetrator\_pronoun\} even being present during the billed treatments.
\end{quote}

\paragraph{Second-hand introduction.}
\begin{quote}
\ttfamily\small
You are a physician at the hospital "Klinik am See". You have already been working with your team
in this hospital for several years. Together with your team, you advise people who are ill on a
daily basis, make diagnoses and carry out treatments.\\[4pt]
The other day you happened to hear about something that happened on your team some time ago:\\[4pt]
\{perpetrator\_ref\} from your team, Dr Schmitt, billed the health insurance company for more
services than were actually provided for some patients. For example, services were billed by
Dr Schmitt without \{perpetrator\_pronoun\} even being present during the billed treatments.
\end{quote}

\paragraph{Condition-specific endings.}
\begin{description}[leftmargin=1.4em]
\item[\textit{no whistleblowing}] (First-hand) ``When \{wb\_ref\} of your colleagues, Dr Bauer,
learned about it, \{wb\_pronoun\} did not pass on any information about it.''\\
(Second-hand) ``You also learned that \{wb\_ref\} of your colleagues, Dr Bauer, learned about it and
subsequently did not pass on any information about it.''
\item[\textit{internal}] (First-hand) ``When \{wb\_ref\} of your colleagues, Dr Bauer, learned about
this, \{wb\_pronoun\} passed on information about it to the hospital's compliance department, which
is responsible for ensuring that the hospital complies with all legal requirements.''\\
(Second-hand) ``You also learned that \{wb\_ref\} of your colleagues, Dr Bauer, learned about it and
subsequently passed on information about it to the hospital's compliance department \ldots''
\item[\textit{external}] (First-hand) ``When \{wb\_ref\} of your colleagues, Dr Bauer, learned about
this, \{wb\_pronoun\} passed on information about it to the relevant authorities.''\\
(Second-hand) ``You also learned that \{wb\_ref\} of your colleagues, Dr Bauer, learned about it and
subsequently passed on information about it to the relevant authorities.''
\item[\textit{public}] (First-hand) ``When \{wb\_ref\} of your colleagues, Dr Bauer, learned about
this, \{wb\_pronoun\} passed on information about it to a newspaper that is distributed throughout
Germany.''\\
(Second-hand) ``You also learned that \{wb\_ref\} of your colleagues, Dr Bauer, learned about it and
subsequently passed on information about it to a newspaper \ldots''
\end{description}

\subsubsection{User prompt (measures and response format)}
After the scenario text, the same user prompt was used across all conditions, models, and
perspectives. Motive and character item order was re-randomized for every query.

\begin{quote}
\ttfamily\small
The scenario:\\[2pt]
\{scenario\_text\}\\[4pt]
As a reminder: Dr Bauer was the person who learned that Dr Schmitt was billing the insurance
company for more services than were actually provided.\\[4pt]
Please rate the extent to which you agree with each statement, using this scale:\\
1 = strongly disagree, 2 = disagree, 3 = rather disagree, 4 = rather agree, 5 = agree,
6 = strongly agree\\[4pt]
Motive items (Dr Bauer wanted to...): \ldots (12 items; Supplementary Table~\ref{tab:S1items})\\[4pt]
Character items (Dr Bauer is...): \ldots (16 items; Supplementary Table~\ref{tab:S1items})\\[4pt]
Respond ONLY with this exact JSON structure (no extra text):\\
\{ "motive\_items": \{ "deontic\_1": <1-6>, ... \}, "character\_items": \{ "honest": <1-6>, ...
\} \}
\end{quote}

\subsubsection{Models, access routes, and sampling parameters}
All models were queried with temperature $t = 0.8$ and each provider's default \texttt{top\_p}.
\textbf{DeepSeek-V3} via the DeepSeek API (\texttt{deepseek-chat}; \texttt{max\_tokens} 600).
\textbf{GPT-4o} via the OpenAI API (\texttt{gpt-4o-2024-08-06}; 600).
\textbf{Gemini-2.5-Pro} via Google's OpenAI-compatible endpoint (\texttt{gemini-2.5-pro}; 4000,
raised because tokens consumed by the internal reasoning pass count against the budget; empty
responses were treated as failures and retried).
\textbf{Llama-3.1-8B-Instruct} via the Hugging Face Inference API (600; serving configuration
determined by the provider).
\textbf{gpt-oss-120b} via the Cerebras API (4000, for the same reason as Gemini-2.5-Pro).

\begin{table}[ht]
\centering
\caption{Full wording of the 12 motive-attribution items and 16 character-judgment items.}
\label{tab:S1items}
\small
\begin{tabular}{@{}ll@{}}
\toprule
Item key & Wording \\
\midrule
deontic\_1 & Dr Bauer wanted to follow a moral obligation. \\
deontic\_2 & Dr Bauer wanted to act in accordance with moral values. \\
deontic\_3 & Dr Bauer wanted to do the right thing. \\
prosocial\_1 & Dr Bauer wanted to help others. \\
prosocial\_2 & Dr Bauer wanted to support others. \\
prosocial\_3 & Dr Bauer wanted to assist others. \\
individualistic\_1 & Dr Bauer wanted to benefit themselves. \\
individualistic\_2 & Dr Bauer wanted to pursue their own interests. \\
individualistic\_3 & Dr Bauer wanted to gain an advantage for themselves. \\
competitive\_1 & Dr Bauer wanted to harm others. \\
competitive\_2 & Dr Bauer wanted to humiliate others. \\
competitive\_3 & Dr Bauer wanted to make others look bad. \\
\midrule
humble, kind, forgiving, giving, helpful, & Dr Bauer is [trait]. \\
grateful, empathetic, cooperative, courageous, & \\
fair, principled, responsible, just, honest, & \\
trustworthy, loyal & \\
\bottomrule
\end{tabular}
\end{table}

\section{Supplementary Results}\label{sec:results}

\subsection{Condition-level descriptive statistics}\label{sec:S2}

\begin{table}[ht]
\centering
\caption{Condition-level means for the two human samples.}
\label{tab:S2human}
\small
\begin{tabular}{@{}llcccccc@{}}
\toprule
Sample & Condition & $n$ & Deontic & Prosocial & Individualistic & Competitive & Character \\
\midrule
First-hand ($N=125$) & No whistleblowing & 30 & 2.37 & 3.24 & 3.80 & 2.06 & 2.70 \\
 & Internal & 33 & 5.62 & 3.38 & 2.44 & 1.98 & 4.34 \\
 & External & 30 & 5.03 & 3.53 & 3.01 & 2.50 & 4.05 \\
 & Public & 32 & 4.08 & 2.80 & 3.56 & 3.68 & 3.38 \\
\addlinespace
Second-hand ($N=742$) & No whistleblowing & 179 & 2.33 & 3.16 & 3.28 & 1.88 & 2.64 \\
 & Internal & 189 & 5.09 & 3.42 & 2.56 & 2.13 & 4.07 \\
 & External & 190 & 4.86 & 3.16 & 2.67 & 2.29 & 3.87 \\
 & Public & 184 & 3.49 & 2.66 & 3.68 & 3.74 & 2.83 \\
\bottomrule
\end{tabular}
\end{table}

\begin{table}[ht]
\centering
\caption{Condition-level means for the five models, first-hand materials
($n = 200$ per condition per model).}
\label{tab:S3fh}
\small
\begin{tabular}{@{}llccccc@{}}
\toprule
Model & Condition & Deontic & Prosocial & Individualistic & Competitive & Character \\
\midrule
DeepSeek-V3 & No whistleblowing & 3.98 & 3.10 & 3.05 & 1.38 & 3.46 \\
 & Internal & 5.80 & 4.86 & 1.90 & 1.09 & 4.70 \\
 & External & 5.76 & 4.87 & 1.90 & 1.10 & 4.68 \\
 & Public & 5.61 & 4.79 & 2.16 & 1.32 & 4.36 \\
\addlinespace
GPT-4o & No whistleblowing & 2.26 & 2.27 & 2.73 & 1.61 & 2.53 \\
 & Internal & 5.93 & 4.88 & 1.94 & 1.28 & 4.96 \\
 & External & 5.94 & 4.93 & 1.95 & 1.33 & 4.92 \\
 & Public & 5.78 & 4.79 & 2.12 & 1.76 & 4.51 \\
\addlinespace
Gemini-2.5-Pro & No whistleblowing & 1.01 & 2.96 & 4.78 & 1.06 & 1.95 \\
 & Internal & 6.00 & 5.08 & 1.94 & 1.23 & 4.46 \\
 & External & 6.00 & 5.15 & 1.85 & 1.24 & 4.29 \\
 & Public & 5.90 & 5.12 & 2.28 & 2.18 & 3.71 \\
\addlinespace
Llama-3.1-8B-Instruct & No whistleblowing & 4.88 & 3.77 & 2.14 & 1.55 & 4.09 \\
 & Internal & 5.79 & 5.27 & 1.50 & 1.31 & 5.08 \\
 & External & 5.82 & 4.96 & 1.66 & 1.49 & 4.98 \\
 & Public & 4.89 & 3.04 & 2.32 & 2.63 & 4.08 \\
\addlinespace
gpt-oss-120b & No whistleblowing & 1.74 & 1.75 & 4.64 & 1.84 & 2.19 \\
 & Internal & 5.75 & 5.33 & 1.46 & 1.33 & 4.91 \\
 & External & 5.76 & 5.33 & 1.57 & 1.37 & 4.94 \\
 & Public & 5.51 & 5.16 & 1.70 & 1.67 & 4.55 \\
\bottomrule
\end{tabular}
\end{table}

\begin{table}[ht]
\centering
\caption{Condition-level means for the five models, second-hand materials
($n = 200$ per condition per model).}
\label{tab:S4sh}
\footnotesize
\setlength{\tabcolsep}{4pt}
\begin{tabular}{@{}llccccc@{}}
\toprule
Model & Condition & Deontic & Prosocial & Individualistic & Competitive & Character \\
\midrule
DeepSeek-V3 & No whistleblowing & 3.81 & 3.06 & 3.37 & 1.42 & 3.08 \\
 & Internal & 5.74 & 4.80 & 1.97 & 1.08 & 4.66 \\
 & External & 5.76 & 4.86 & 1.97 & 1.09 & 4.66 \\
 & Public & 5.58 & 4.71 & 2.25 & 1.40 & 4.36 \\
\addlinespace
GPT-4o & No whistleblowing & 2.31 & 2.15 & 2.90 & 1.76 & 2.46 \\
 & Internal & 5.93 & 4.91 & 1.95 & 1.24 & 5.02 \\
 & External & 5.93 & 4.89 & 1.99 & 1.39 & 4.94 \\
 & Public & 5.70 & 4.75 & 2.17 & 1.77 & 4.50 \\
\addlinespace
Gemini-2.5-Pro & No whistleblowing & 1.01 & 2.70 & 4.84 & 1.06 & 1.90 \\
 & Internal & 6.00 & 5.10 & 1.85 & 1.16 & 4.39 \\
 & External & 6.00 & 5.16 & 1.82 & 1.25 & 4.29 \\
 & Public & 5.94 & 5.12 & 2.32 & 2.21 & 3.70 \\
\addlinespace
Llama-3.1-8B-Instruct & No whistleblowing & 4.78 & 3.16 & 2.43 & 1.60 & 4.04 \\
 & Internal & 5.72 & 5.25 & 1.60 & 1.38 & 5.08 \\
 & External & 5.75 & 5.10 & 1.70 & 1.42 & 5.12 \\
 & Public & 5.41 & 3.61 & 1.98 & 1.80 & 4.56 \\
\addlinespace
gpt-oss-120b & No whistleblowing & 1.78 & 1.81 & 4.66 & 1.86 & 2.27 \\
 & Internal & 5.70 & 5.21 & 1.58 & 1.42 & 4.87 \\
 & External & 5.70 & 5.24 & 1.56 & 1.39 & 4.88 \\
 & Public & 5.54 & 5.11 & 1.77 & 1.73 & 4.52 \\
\bottomrule
\end{tabular}
\end{table}

\subsection{\rev{Precision of the outcome-level correlations}}\label{sec:S3}

\rev{Each correlation in the main text is computed over four condition means, so its sampling
variability is substantial. Supplementary Table~\ref{tab:S5boot} reports percentile bootstrap intervals from
2,000 resamples of respondents within condition. The character-judgment and deontic correlations
are precisely estimated; the prosocial correlation is not, with intervals spanning most of the
possible range. The main text therefore rests on condition contrasts and absolute levels rather
than on these correlations.}

\begin{table}[ht]
\centering
\caption{\rev{Correlation between human and model condition means, with 95\% percentile bootstrap
intervals (2,000 resamples).}}
\label{tab:S5boot}
\scriptsize
\setlength{\tabcolsep}{2.5pt}
\begin{tabular}{@{}llccccc@{}}
\toprule
Comparison & Model & Character & Deontic & Prosocial & Individualistic & Competitive \\
\midrule
First-hand & DeepSeek-V3 & 0.95 [0.81, 0.99] & 0.93 [0.85, 0.98] & 0.03 [$-$0.70, 0.65] & 0.79 [0.49, 0.94] & 0.34 [0.01, 0.63] \\
 & GPT-4o & 0.92 [0.78, 0.99] & 0.91 [0.82, 0.96] & 0.04 [$-$0.70, 0.64] & 0.79 [0.50, 0.94] & 0.70 [0.46, 0.85] \\
 & Gemini-2.5-Pro & 0.96 [0.84, 1.00] & 0.90 [0.81, 0.96] & $-$0.00 [$-$0.67, 0.64] & 0.73 [0.42, 0.91] & 0.96 [0.85, 1.00] \\
 & Llama-3.1-8B-Instruct & 0.92 [0.73, 0.99] & 0.86 [0.69, 0.96] & 0.90 [0.08, 0.99] & 0.92 [0.62, 0.99] & 0.96 [0.82, 1.00] \\
 & gpt-oss-120b & 0.90 [0.74, 0.98] & 0.92 [0.83, 0.97] & 0.04 [$-$0.68, 0.70] & 0.70 [0.41, 0.89] & 0.23 [$-$0.05, 0.47] \\
\addlinespace
Second-hand & DeepSeek-V3 & 0.78 [0.71, 0.85] & 0.88 [0.83, 0.92] & $-$0.07 [$-$0.37, 0.22] & 0.48 [0.31, 0.63] & 0.37 [0.14, 0.59] \\
 & GPT-4o & 0.79 [0.72, 0.85] & 0.87 [0.82, 0.91] & $-$0.07 [$-$0.36, 0.23] & 0.49 [0.33, 0.65] & 0.42 [0.29, 0.55] \\
 & Gemini-2.5-Pro & 0.83 [0.77, 0.89] & 0.84 [0.79, 0.89] & $-$0.13 [$-$0.39, 0.16] & 0.44 [0.28, 0.60] & 1.00 [0.98, 1.00] \\
 & Llama-3.1-8B-Instruct & 0.94 [0.88, 0.98] & 0.97 [0.92, 0.99] & 0.59 [0.31, 0.81] & 0.68 [0.45, 0.87] & 0.76 [0.40, 0.94] \\
 & gpt-oss-120b & 0.75 [0.67, 0.82] & 0.86 [0.81, 0.90] & $-$0.09 [$-$0.36, 0.20] & 0.35 [0.18, 0.52] & 0.19 [$-$0.02, 0.38] \\
\bottomrule
\end{tabular}
\end{table}

\begin{table}[ht]
\centering
\caption{Tucker's congruence coefficient $\phi$ between the human and model 20-cell profiles, with
reference baselines. Because all cell means lie between 1 and 6, $\phi$ is high for essentially any
pair of profiles, which is why it is not used in the main text.}
\label{tab:S6phi}
\small
\begin{tabular}{@{}llc@{}}
\toprule
Comparison & Model or baseline & $\phi$ \\
\midrule
First-hand & DeepSeek-V3 & .952 \\
 & GPT-4o & .955 \\
 & Gemini-2.5-Pro & .953 \\
 & Llama-3.1-8B-Instruct & .956 \\
 & gpt-oss-120b & .943 \\
 & \textit{Baseline: constant profile (all cells 3.5)} & .964 \\
 & \textit{Baseline: no condition effect (human grand mean per outcome)} & .978 \\
 & \textit{Baseline: human conditions randomly permuted (mean of 23)} & .954 \\
\addlinespace
Second-hand & DeepSeek-V3 & .948 \\
 & GPT-4o & .949 \\
 & Gemini-2.5-Pro & .942 \\
 & Llama-3.1-8B-Instruct & .944 \\
 & gpt-oss-120b & .936 \\
 & \textit{Baseline: constant profile (all cells 3.5)} & .967 \\
 & \textit{Baseline: no condition effect (human grand mean per outcome)} & .977 \\
 & \textit{Baseline: human conditions randomly permuted (mean of 23)} & .953 \\
\bottomrule
\end{tabular}
\end{table}

\subsection{\rev{Measurement properties and response diversity}}\label{sec:S4}

\begin{table}[ht]
\centering
\caption{\rev{Internal consistency (Cronbach's $\alpha$), proportion of responses with all three
deontic items at the scale maximum, and the range of distinct 28-item response vectors among the
200 responses in each cell.}}
\label{tab:S7rel}
\scriptsize
\setlength{\tabcolsep}{3.5pt}
\begin{tabular}{@{}lccccccc@{}}
\toprule
Respondent & Deontic & Prosocial & Individ. & Compet. & Character & Deontic ceiling & Distinct vectors \\
\midrule
Human (first-hand) & 0.97 & 0.89 & 0.83 & 0.88 & 0.95 & --- & --- \\
Human (second-hand) & 0.95 & 0.89 & 0.82 & 0.91 & 0.95 & --- & --- \\
\addlinespace
DeepSeek-V3 & 0.99 & 0.99 & 0.97 & 0.85 & 0.94 & 48\% & 151--197 \\
GPT-4o & 1.00 & 0.99 & 0.96 & 0.82 & 0.97 & 62\% & 155--187 \\
Gemini-2.5-Pro & 1.00 & 0.98 & 0.98 & 0.87 & 0.90 & 71\% & 195--200 \\
Llama-3.1-8B-Instruct & 0.87 & 0.93 & 0.94 & 0.90 & 0.93 & 47\% & 200--200 \\
gpt-oss-120b & 1.00 & 1.00 & 0.99 & 0.90 & 0.98 & 45\% & 173--198 \\
\bottomrule
\end{tabular}
\end{table}

\subsection{Between-sample differences in the human data}\label{sec:S5}

\begin{table}[ht]
\centering
\caption{Differences between the second-hand and first-hand human samples (scale
points). Main effect: perspective coefficient controlling for disclosure condition (HC3).
Interaction: joint test of the condition-by-sample terms (3 d.f.). Per-condition columns give the
mean difference with Cohen's $d$ in parentheses.}
\label{tab:S8human}
\small
\begin{tabular}{@{}lcccccc@{}}
\toprule
 & \multicolumn{2}{c}{Main effect} & \multicolumn{4}{c}{Per condition: difference ($d$)} \\
\cmidrule(lr){2-3}\cmidrule(lr){4-7}
Dimension & coef. & $P$ & No WB & Internal & External & Public \\
\midrule
Deontic & $-0.34$ & .002 & $-0.04$ ($-0.03$) & $-0.52$ ($-0.65$) & $-0.18$ ($-0.16$) & $-0.59$ ($-0.41$) \\
Prosocial & $-0.13$ & .269 & $-0.08$ ($-0.08$) & $+0.04$ ($+0.03$) & $-0.37$ ($-0.31$) & $-0.14$ ($-0.12$) \\
Individualistic & $-0.15$ & .174 & $-0.52$ ($-0.48$) & $+0.11$ ($+0.12$) & $-0.34$ ($-0.31$) & $+0.12$ ($+0.10$) \\
Competitive & $-0.04$ & .702 & $-0.18$ ($-0.22$) & $+0.15$ ($+0.18$) & $-0.21$ ($-0.20$) & $+0.07$ ($+0.05$) \\
\bottomrule
\end{tabular}
\end{table}

\subsection{Sensitivity analyses}\label{sec:S6}

In each of 300 draws, every model pool was subsampled without replacement to the human sample size
in each perspective-by-condition cell, and the tests were re-estimated. Supplementary Table~\ref{tab:S9sens}
reports the proportion of draws significant at $P < 0.05$. This checks whether a result depends on
the model pools being larger than the human samples; it is not a power analysis.
\rev{Gemini-2.5-Pro's deontic composite has zero variance in the internal and external conditions.
Re-estimating its attribution--judgment coefficients with those conditions excluded changes which
of its interactions reach significance (deontic becomes significant, $P = 0.015$; competitive does
not, $P = 0.714$), so that model's structure results are treated with more caution in the main
text.}

\begin{table}[ht]
\centering
\caption{Proportion of 300 size-matched subsamples in which the between-sample joint test and the
attribution--judgment interaction were significant at $P < 0.05$.}
\label{tab:S9sens}
\small
\begin{tabular}{@{}lcccccccc@{}}
\toprule
 & \multicolumn{2}{c}{Deontic} & \multicolumn{2}{c}{Prosocial} & \multicolumn{2}{c}{Individ.} & \multicolumn{2}{c}{Compet.} \\
\cmidrule(lr){2-3}\cmidrule(lr){4-5}\cmidrule(lr){6-7}\cmidrule(lr){8-9}
Model & Sample & Struct. & Sample & Struct. & Sample & Struct. & Sample & Struct. \\
\midrule
DeepSeek-V3 & 1.00 & 1.00 & 0.00 & 0.85 & 0.67 & 0.00 & 0.00 & 1.00 \\
GPT-4o & 1.00 & 0.01 & 0.00 & 1.00 & 0.80 & 0.03 & 0.10 & 0.59 \\
Gemini-2.5-Pro & 1.00 & 0.00 & 0.00 & 1.00 & 0.59 & 0.00 & 0.00 & 1.00 \\
Llama-3.1-8B-Instruct & 1.00 & 0.98 & 0.32 & 1.00 & 0.52 & 0.00 & 0.35 & 0.75 \\
gpt-oss-120b & 0.99 & 0.91 & 0.00 & 0.00 & 0.23 & 0.04 & 0.00 & 0.99 \\
\bottomrule
\end{tabular}
\end{table}

\subsection{\rev{Persona sensitivity and persona-matched estimates}}\label{sec:S7}

\rev{Personas were randomly sampled within each model's generation pipeline, so the association
between persona attributes and ratings is estimated without confounding. Supplementary Table~\ref{tab:S10persona}
expresses the age coefficient as the change implied across the 15.7-year difference in mean age
between the two human samples, and gives the student-status coefficients. Both are small.}

\begin{table}[ht]
\centering
\caption{\rev{Association between persona attributes and model ratings, controlling for disclosure
condition and framing. Age columns give the change implied across a 15.7-year age difference;
student columns give the coefficient for student status.}}
\label{tab:S10persona}
\small
\begin{tabular}{@{}lcccccccc@{}}
\toprule
 & \multicolumn{4}{c}{Implied change across 15.7 years of age} & \multicolumn{4}{c}{Student status} \\
\cmidrule(lr){2-5}\cmidrule(lr){6-9}
Model & Deon. & Prosoc. & Indiv. & Compet. & Deon. & Prosoc. & Indiv. & Compet. \\
\midrule
DeepSeek-V3 & $+0.09$ & $+0.08$ & $-0.06$ & $-0.03$ & $-0.06$ & $-0.03$ & $-0.01$ & $+0.03$ \\
GPT-4o & $-0.01$ & $-0.00$ & $-0.04$ & $+0.01$ & $+0.00$ & $+0.02$ & $+0.00$ & $+0.04$ \\
Gemini-2.5-Pro & $+0.00$ & $-0.02$ & $-0.09$ & $-0.03$ & $+0.01$ & $-0.01$ & $-0.01$ & $-0.04$ \\
Llama-3.1-8B-Instruct & $+0.07$ & $+0.22$ & $-0.04$ & $+0.01$ & $+0.12$ & $-0.06$ & $-0.18$ & $-0.13$ \\
gpt-oss-120b & $-0.01$ & $-0.02$ & $+0.01$ & $+0.01$ & $-0.01$ & $+0.03$ & $+0.02$ & $+0.02$ \\
\bottomrule
\end{tabular}
\end{table}

\rev{Supplementary Table~\ref{tab:S11matched} reports the model between-sample estimates three ways: as run
(both the framing sentence and the matched persona roster differ), adjusting for persona age,
gender and student status, and within a persona-matched subsample retaining every age-band by
gender by student-status by condition by model cell present under both framings (6,777 of 8,000
responses per model set). The estimates are unchanged, so the null is not produced by the persona
rosters. The final block reports two one-sided tests against the corresponding human difference as
the equivalence bound. For the competitive dimension the human difference is itself near zero
($-0.04$), so the bound is too narrow for the test to be informative, and we do not interpret it.}

\begin{table}[ht]
\centering
\caption{\rev{Model between-sample estimates (second-hand minus first-hand, scale points),
controlling for disclosure condition, and equivalence tests against the human difference.}}
\label{tab:S11matched}
\footnotesize
\setlength{\tabcolsep}{4pt}
\begin{tabular}{@{}lcccc@{}}
\toprule
\multicolumn{5}{@{}l}{\textit{Deontic: as run / persona-adjusted / persona-matched [95\% CI]}} \\
\midrule
DeepSeek-V3 & $-0.064$ & $-0.167$ & $-0.087$ & $[-0.17, -0.00]$ \\
GPT-4o & $-0.006$ & $-0.000$ & $-0.003$ & $[-0.04, +0.04]$ \\
Gemini-2.5-Pro & $+0.012$ & $+0.017$ & $+0.011$ & $[-0.00, +0.02]$ \\
Llama-3.1-8B-Instruct & $+0.073$ & $+0.042$ & $+0.060$ & $[-0.05, +0.17]$ \\
gpt-oss-120b & $-0.009$ & $-0.003$ & $-0.015$ & $[-0.06, +0.03]$ \\
Pooled & $+0.001$ & --- & $-0.023$ & $[-0.07, +0.02]$ \\
\textit{Human between-sample difference} & \multicolumn{4}{c}{$-0.341$ $[-0.55, -0.13]$, $P = .002$} \\
\midrule
\multicolumn{5}{@{}l}{\textit{Persona-matched estimates, all dimensions}} \\
\midrule
 & Deontic & Prosocial & Individualistic & Competitive \\
DeepSeek-V3 & $-0.087$ & $-0.059$ & $+0.164$ & $+0.034$ \\
GPT-4o & $-0.003$ & $-0.049$ & $+0.079$ & $+0.041$ \\
Gemini-2.5-Pro & $+0.011$ & $-0.061$ & $+0.022$ & $-0.010$ \\
Llama-3.1-8B-Instruct & $+0.060$ & $+0.006$ & $+0.002$ & $-0.228$ \\
gpt-oss-120b & $-0.015$ & $-0.042$ & $+0.061$ & $+0.054$ \\
\midrule
\multicolumn{5}{@{}l}{\textit{Equivalence tests ($P_{\mathrm{TOST}}$, bound = human difference)}} \\
\midrule
 & \multicolumn{4}{c}{DeepSeek-V3 / GPT-4o / Gemini-2.5-Pro / Llama-3.1-8B-Instruct / gpt-oss-120b} \\
Deontic (bound $0.341$) & \multicolumn{4}{c}{$<$.001 / $<$.001 / $<$.001 / $<$.001 / $<$.001} \\
Prosocial (bound $0.134$) & \multicolumn{4}{c}{.015 / $<$.001 / .025 / .069 / $<$.001} \\
Individualistic (bound $0.147$) & \multicolumn{4}{c}{.682 / .001 / $<$.001 / .010 / $<$.001} \\
Competitive (bound $0.037$) & \multicolumn{4}{c}{not informative (see text)} \\
\bottomrule
\end{tabular}
\end{table}

\subsection{\rev{Weighting, a priori equivalence bounds, and random-effects pooling}}\label{sec:S9}

\rev{Supplementary Table~\ref{tab:S12robust} reports three additional analyses of the model-side estimates.
The first panel gives covariate balance between the two specifications before and after stabilized
inverse-probability weighting, expressed as standardized mean differences, together with the
weighted estimate for duty-based attribution. Weighting substantially reduces the imbalance but
does not eliminate it (age falls from about 1.2 to about 0.3), so the weighted estimates are
reported alongside the persona-matched estimates in Supplementary Table~\ref{tab:S11matched} rather than in place
of them. All three approaches, adjustment, matching, and weighting, place every model far closer to
zero than the human difference of $-0.34$.}

\rev{The second panel repeats the equivalence tests against an a priori bound of 0.20 scale points,
chosen independently of the data as the smallest difference of interest on a six-point scale. This
bound gives the same conclusion for duty-based attribution as the data-derived bound and, unlike
it, is informative for the competitive dimension, where the human difference is itself close to
zero.}

\rev{The third panel pools the five model-level estimates with a DerSimonian-Laird random-effects
model, treating model as a random factor. The prediction interval indicates the range within which
the estimate for a further model of the same kind would be expected to fall. For duty-based
attribution that interval, $[-0.08, +0.07]$, excludes the human difference by a wide margin.}

\begin{table}[ht]
\centering
\caption{\rev{Robustness of the model-side between-sample estimates.}}
\label{tab:S12robust}
\footnotesize
\begin{tabular}{@{}lccccccc@{}}
\toprule
\multicolumn{8}{@{}l}{\textit{Panel A. Covariate balance (standardized mean difference) and inverse-probability-weighted estimate}} \\
\midrule
 & \multicolumn{2}{c}{Age} & \multicolumn{2}{c}{Student} & \multicolumn{2}{c}{Female} & Weighted \\
\cmidrule(lr){2-3}\cmidrule(lr){4-5}\cmidrule(lr){6-7}
Model & before & after & before & after & before & after & deontic estimate \\
\midrule
DeepSeek-V3 & $+1.20$ & $+0.28$ & $-0.72$ & $-0.14$ & $-0.28$ & $-0.06$ & $-0.149$ $[-0.24, -0.06]$ \\
GPT-4o & $+1.20$ & $+0.27$ & $-0.73$ & $-0.14$ & $-0.28$ & $-0.06$ & $-0.004$ $[-0.07, +0.06]$ \\
Gemini-2.5-Pro & $+1.16$ & $+0.30$ & $-0.68$ & $-0.18$ & $-0.30$ & $-0.09$ & $+0.012$ $[+0.00, +0.02]$ \\
Llama-3.1-8B-Instruct & $+1.33$ & $+0.37$ & $-0.70$ & $-0.17$ & $-0.30$ & $-0.07$ & $+0.038$ $[-0.07, +0.15]$ \\
gpt-oss-120b & $+1.07$ & $+0.28$ & $-0.69$ & $-0.18$ & $-0.30$ & $-0.08$ & $+0.002$ $[-0.05, +0.05]$ \\
\addlinespace
\multicolumn{8}{@{}l}{\textit{Panel B. Equivalence against an a priori bound of $\pm 0.20$ scale points ($P_{\mathrm{TOST}}$)}} \\
\midrule
Dimension & DeepSeek-V3 & GPT-4o & Gemini & Llama & gpt-oss & \multicolumn{2}{c}{Equivalent} \\
\midrule
Deontic & .004 & $<$.001 & $<$.001 & .006 & $<$.001 & \multicolumn{2}{c}{5/5} \\
Prosocial & $<$.001 & $<$.001 & $<$.001 & .012 & $<$.001 & \multicolumn{2}{c}{5/5} \\
Individualistic & .163 & $<$.001 & $<$.001 & .001 & $<$.001 & \multicolumn{2}{c}{4/5} \\
Competitive & $<$.001 & $<$.001 & $<$.001 & .694 & $<$.001 & \multicolumn{2}{c}{4/5} \\
\addlinespace
\multicolumn{8}{@{}l}{\textit{Panel C. Random-effects pooling of the five model estimates}} \\
\midrule
Dimension & \multicolumn{2}{c}{Pooled [95\% CI]} & $\tau^2$ & $I^2$ & \multicolumn{2}{c}{Prediction interval} & Human \\
\midrule
Deontic & \multicolumn{2}{c}{$-0.003$ $[-0.03, +0.02]$} & 0.0004 & 47\% & \multicolumn{2}{c}{$[-0.08, +0.07]$} & $-0.341$ \\
Prosocial & \multicolumn{2}{c}{$-0.048$ $[-0.07, -0.02]$} & 0.0000 & 0\% & \multicolumn{2}{c}{$[-0.09, -0.01]$} & $-0.134$ \\
Individualistic & \multicolumn{2}{c}{$+0.070$ $[+0.02, +0.12]$} & 0.0016 & 65\% & \multicolumn{2}{c}{$[-0.08, +0.22]$} & $-0.147$ \\
Competitive & \multicolumn{2}{c}{$-0.005$ $[-0.06, +0.05]$} & 0.0034 & 84\% & \multicolumn{2}{c}{$[-0.21, +0.20]$} & $-0.037$ \\
\bottomrule
\end{tabular}
\end{table}

\clearpage
\section{Supplementary Figures}\label{sec:figures}

\begin{figure}[ht]
\centering
\includegraphics[width=\linewidth]{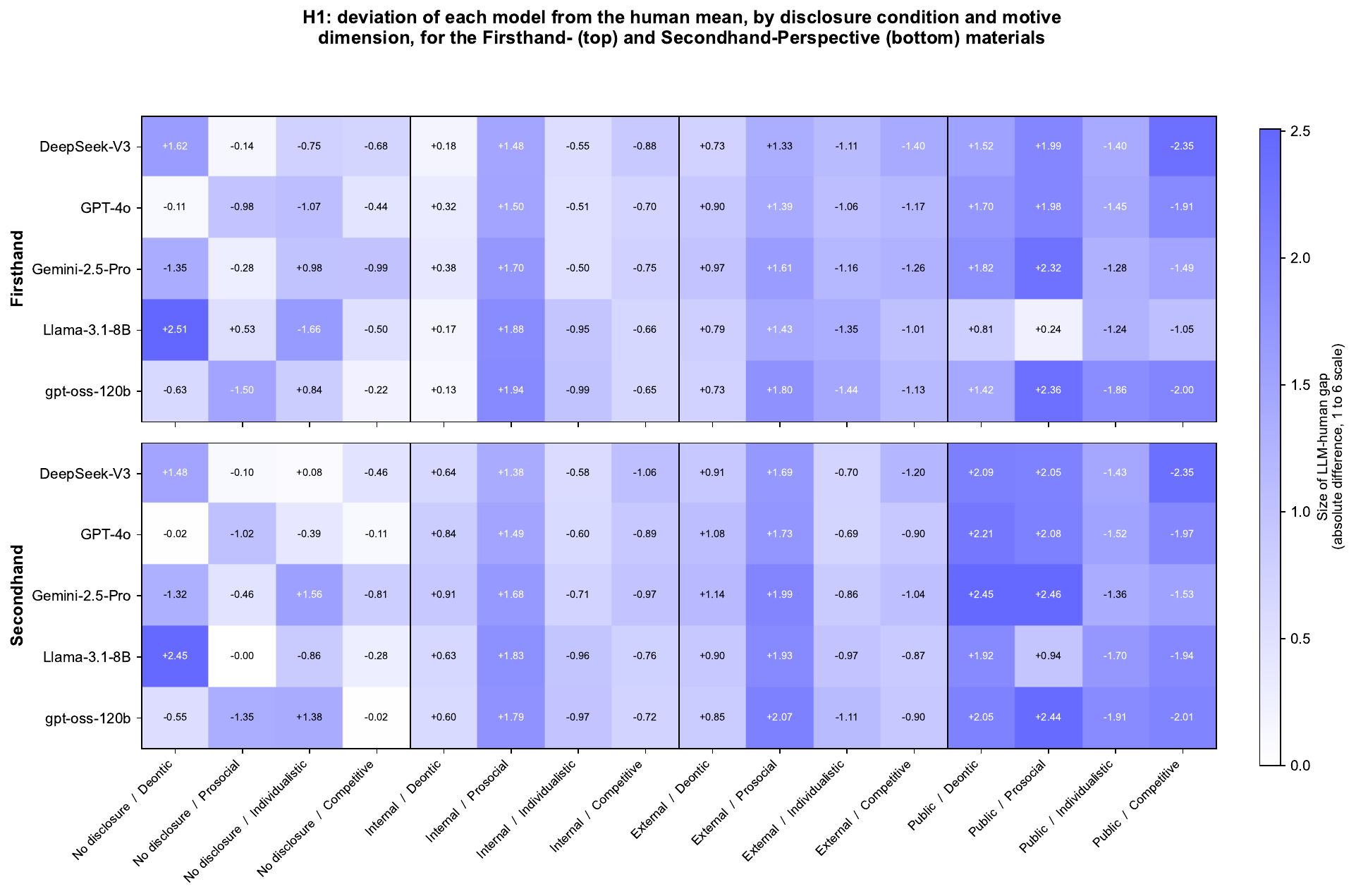}
\caption{Cell-by-cell deviation of each model from the human mean (model mean minus human mean, on
the 1--6 rating scale), for all 16 condition-by-dimension combinations, for the first-hand (top block) and second-hand (bottom block) materials. Rows are the five models; columns
are the four disclosure conditions, each subdivided into the four motive dimensions. The two blocks
are visually very similar, which is itself informative: the two sets of materials produce
essentially the same human--model deviation pattern.}
\label{fig:S1heatmap}
\end{figure}

\begin{figure}[ht]
\centering
\includegraphics[width=0.92\linewidth]{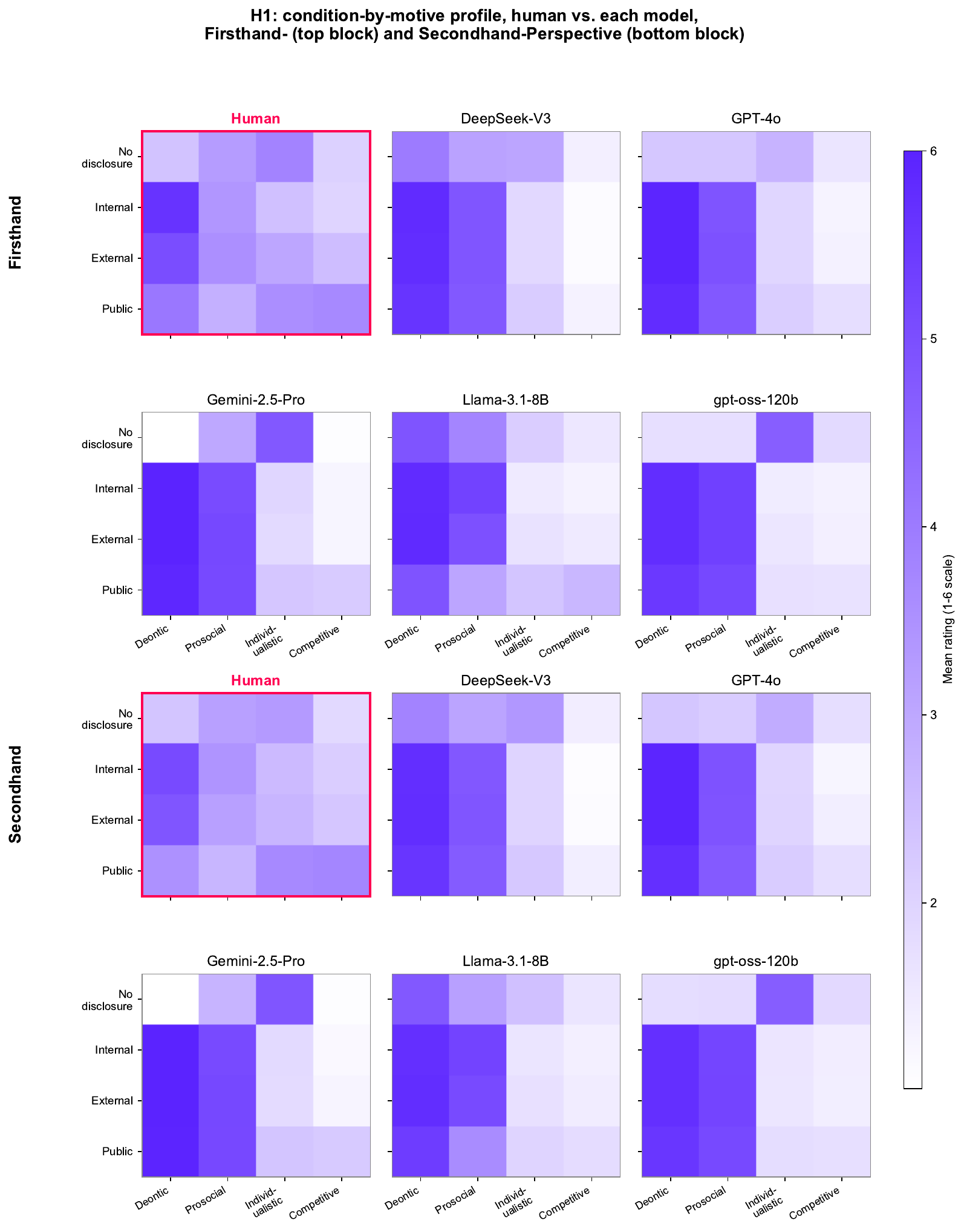}
\caption{Condition-by-motive-dimension profile (mean rating, 1--6 scale) for the human sample and
each of the five models, for the first-hand (top block) and second-hand (bottom block)
materials, on a shared colour scale.}
\label{fig:S2smallmult}
\end{figure}
